\def\year{2022}\relax
\documentclass[letterpaper]{article} 
\usepackage{aaai22}  
\usepackage{times}  
\usepackage{helvet}  
\usepackage{courier}  
\usepackage[hyphens]{url}  
\usepackage{graphicx} 
\usepackage{natbib}  
\usepackage{caption}
\usepackage{amsmath}
\usepackage{multirow}
\DeclareCaptionStyle{ruled}{labelfont=normalfont,labelsep=colon,strut=off} 
\usepackage{algorithm}
\usepackage{algorithmic}
\usepackage{subfigure}
\usepackage{wrapfig}

\usepackage{newfloat}
\usepackage{listings}
\floatstyle{ruled}
\newfloat{listing}{tb}{lst}{}
\floatname{listing}{Listing}
\newcommand{\softmax}{\mathrm{softmax}}
\newcommand{\softplus}{\mathrm{softplus}}
\title{Mixture Density Networks for Classification with an Application to Product Bundling}
\author{Narendhar Gugulothu\textsuperscript{\rm 1}, Sanjay P. Bhat\textsuperscript{\rm 1}, Tejas Bodas \textsuperscript{\rm 2}}
\affiliations{\textsuperscript{\rm 1}TCS Research, Hyderabad, India\\
\textsuperscript{\rm 2} Computer Systems Group, IIIT-Hyderabad, India\\
\{narendhar.g, sanjay.bhat\}@tcs.com, tejas.bodas@iiit.ac.in}

\usepackage{bibentry}

\begin{document}

\maketitle

\begin{abstract}
While mixture density networks (MDNs) have been extensively used for regression tasks, they have not been used much for classification tasks. One reason for this is that the usability of MDNs for
classification is not clear and straightforward. In this paper, we propose
two MDN-based models for classification tasks. Both models fit mixtures of Gaussians to the the data and use the fitted distributions
to classify a given sample by evaluating the learnt cumulative distribution function for the given input features. While the proposed MDN-based models perform slightly better than, or on par with, five baseline classification models
on three publicly available datasets, the real utility of our models comes out through a real-world
product bundling application. Specifically, we use our MDN-based models to
learn the willingness-to-pay (WTP) distributions for two products from
synthetic sales data of the individual products. The Gaussian
mixture representation of the learnt WTP distributions is then exploited to obtain the WTP distribution of the bundle consisting of both the products. The proposed MDN-based models are able to approximate the true WTP distributions of both products and the bundle well.

\end{abstract}

\section{Introduction}\label{sec:intro}
Mixture density networks~\cite{bishop1994mixture} have been used in regression tasks due to their direct applicability to regression and superiority in modelling the intrinsic multi-modality~\cite{nilsson2020prediction,sedlmeier2021quantifying} of the data by leveraging mixtures of distributions.
Despite their superiority, MDNs have not been used for classification tasks as the use of the MDN parameters for classification is not straightforward and there is no existing approach for doing this.
However, there are certain applications in which learning the parameters of distribution is very crucial along with classification and one such application is product bundling~\cite{hanson1990optimal}.
Product bundling is a strategy of selling two or more products together often at a discounted price.
Of late, product bundling has gained traction and is being used in many fields including retail~\cite{bhargava2012retailer}, music~\cite{cabral2018mixed}, services~\cite{guidon2020transportation}, food~\cite {carroll2018food}, e-commerce~\cite{beheshtian2018novel}, airlines~\cite{wang2020airline} and tourism~\cite{garrod2017collaborative}.

In product bundling, the availability of bundle-level sales data is very scarce making actionable insights difficult to obtain. Choice modelling theory provides a means to use product-level sales data to overcome bundle-data scarcity by training product-level classification models and using them to predict bundle sales~\cite{hensher2018applied,Train09}. Choice modelling theory assumes that a customer purchases a product if her willingness-to-pay for that product exceeds the offer price. To leverage choice modeling theory, we model the WTP as a random variable, and train a classification model with product-level sales data such that the model learns the parameters of the WTP distribution while learning to classify the input sample as a sale or no-sale sample. Here, the product-level sales data is in the form of indicator variables parameterized by the offer price of the product, with the indicator variable taking the value 1 when the product is sold at the offer price and 0 otherwise. Choice modelling theory predicts that a customer's WTP for any given bundle is the sum of her WTP for each of the items contained in that bundle. This opens up the possibility of using learnt product-level WTP distributions to construct the WTP distribution for any specified bundle. The key to doing this tractably is to use a parameterized family of distributions which is rich enough to closely approximate arbitrary WTP distributions. Additionally, the family should be such that the parameters of the WTP distribution of a bundle should be easily computable from the parameters of the WTP distributions of the products constituting that bundle. It is also to be noted that the product-level sales data is multi-modal in nature due to the effect of external factors such as seasonality, holidays, festivals, reviews,
discounts, promotions, and categorization of customers~\cite{alzate2021online,roy2017effect,weng2013consumers}. Hence, it is important that the family of distributions chosen to model the product-level data should be able to capture any multi-modality that may be present in the
sales data.

To achieve the above-mentioned desiderata, we propose two variants of mixture density networks based classification models (MDN-C). Mixture density networks provide the basis for a popular machine learning technique used to capture the multi-modality present in the data and to learn the distribution parameters~\cite{bishop1994mixture,gugulothu2019sparse,nilsson2020prediction,sedlmeier2021quantifying}.
To this end, we pass the input samples through a mixture density network to learn the distribution parameters. 
The learnt parameters of the distribution are then used to evaluate the cumulative distribution function (CDF) at each of the input feature values in the first variant. 
The evaluated CDF values are then fed to
a softmax layer with LASSO penalty~\cite{gugulothu2018sparse,gugulothu2019load,gugulothu2019sparse,tibshirani1996regression} on it's weights to predict class scores. 
The LASSO penalty helps to retain the influence of the CDF values of only the important mixture components on the predicted class scores.
 In the second variant, the CDF values are evaluated with the learnt latent features instead of the original features. The latent features are learnt by passing the inputs through a feed-forward layer having units equal to the number of classes in the dataset. This is done to reduce the computational cost in CDF evaluation when the input dimensions are large. 
 The evaluated CDF values are then normalized to predict class scores. 
The proposed models are trained in an end-to-end manner by calculating the cross-entropy loss between the class labels and predicted class scores. 
The performance of the proposed MDN based classification models are tested on three publicly available classification datasets. 
We used accuracy, precision, recall, and F1-score as quantitative metrics to evaluate the proposed models. 
We also show the efficacy of the proposed MDN-based classification models on product bundling application, in which the first proposed variant of our MDN-based classification model with minor modifications, trained using product-level sales data, is used in learning the bundle WTP distribution.
The proposed MDN-based classification models have the following features:
\begin{itemize}
    \item Learns the parameters of the distribution by approximating the distribution underlying the data.
    \item Performs on par with or better than five baseline classification techniques.
    \item Estimates bundle willingness-to-pay distribution from product-level sales data. 
    \item Easy to use with powerful deep learning techniques such as recurrent neural networks (RNN), convolutional neural networks (CNN) and transformer networks. 
\end{itemize}

\section{Related Work}\label{sec:related work}
Mixture density networks have gained popularity in recent years due to their superiority in modelling multi-modality in the data~\cite{nilsson2020prediction,sedlmeier2021quantifying}cand learning the distribution parameters.  
MDNs have been combined with powerful deep learning techniques such as RNNs, CNNs, and transformer, 
and used in tasks like forecasting~\cite{gugulothu2019sparse}, motion prediction~\cite{dipietro2018unsupervised}, anomaly detection~\cite{guo2018multidimensional}, volatility prediction~\cite{schittenkopf1998volatility}, sign language production~\cite{saunders2021continuous} and knowledge graphs~\cite{errica2021graph}. 
Despite their popularity, MDNs have not been used for classification tasks as the usability is not straight forward.
To the best of our knowledge, our work is the first instance of using cumulative distribution function values of the distributions learnt using MDNs for classification task.

Product bundling is a common revenue management technique where the idea is to offer different products for sale as part of a bundle with the idea of increasing revenue~\cite{Fuerderer13}. Given a particular bundle of products, one is often interested in identifying a revenue-optimal selling price for the bundle.
Apart from pricing, one is also interested in selecting the right combination of bundles to offer for sale. One of the earliest work to consider a joint bundle optimization and pricing problem is~\cite{Bitran07}. Here, a customer's random choice for bundle is modeled using the multinomial logit choice model~\cite{hensher2018applied,Perez16,Train09}. 
A drawback with such models is that the customer's WTP for a bundle is captured by sophisticated random variables such as Gumbel or generalized extreme value, so that the optimization problem is amenable to analysis. In reality, there is no evidence to suggest that the WTP distribution should be restricted to such variables and, in fact, must be discovered or learnt from the underlying data. Our work focuses on this aspect of learning the bundle WTP from existing product-level data using MDNs and using the learnt distribution for the corresponding bundle price optimization.   

\section{Mixture Density Networks for Classification}\label{sec:MDN-C}
\begin{figure*}[th]
	\subfigure[\label{fig:MDN-Var5}MDN-C1]{\includegraphics[trim={0cm 0cm 0.5cm 0cm},clip,width=0.52\linewidth]{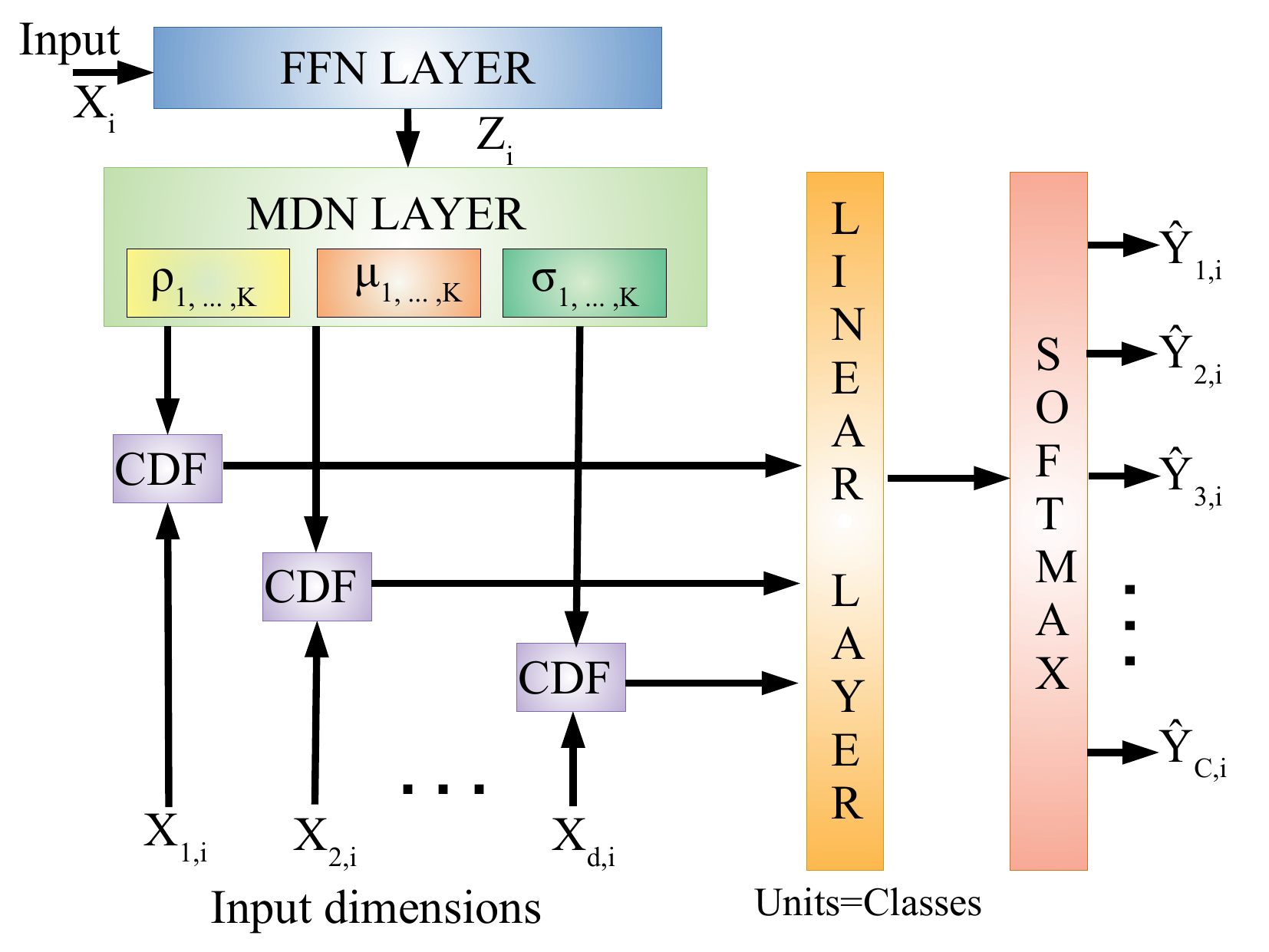}}
	\subfigure[\label{fig:MDN-Var2}MDN-C2 ]{\includegraphics[trim={0cm 0cm 0cm 0cm},clip,width=0.52\linewidth]{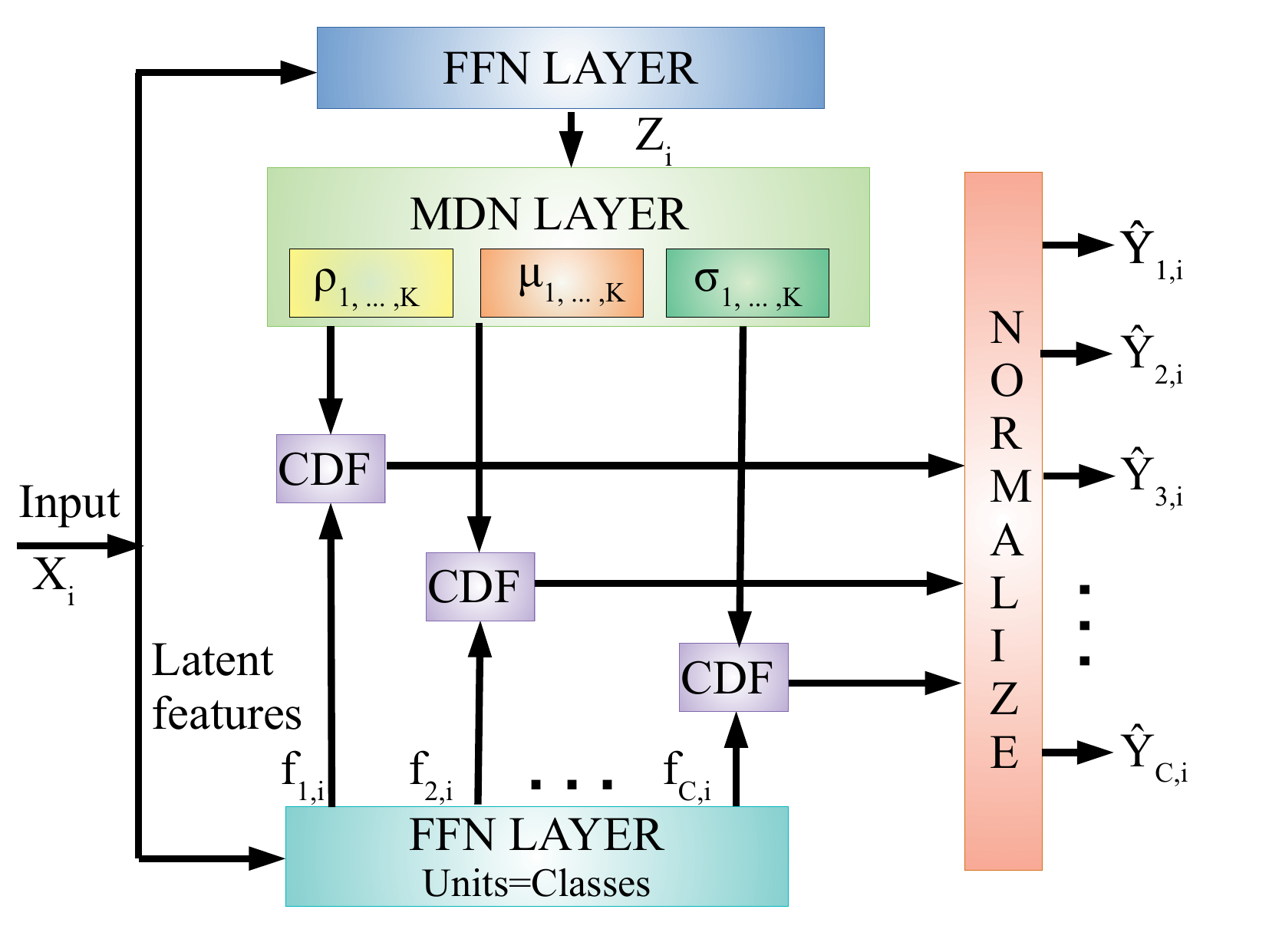}}
	\caption{\label{fig:RMDN} Proposed MDN-based classification models.}
\end{figure*}

In this section, we introduce two variants of MDN architectures, namely, \emph{MDN-C1} and \emph{MDN-C2}, that can be used in classification tasks.
The MDN-based classification models that we present in this exposition are schematically depicted in Figures~\ref{fig:MDN-Var5} and~\ref{fig:MDN-Var2}.
Let $X_i$ denote a $d$-dimensional $i$th input sample. 
The objective of the classification model is to predict the class label given an input sample. In other words, the classification model must predict class scores $\hat{Y}^i_{1},\cdots,\hat{Y}^i_{C}$ given an input sample $X_i$, where $C$ is the number of classes. 

The MDN models use a mixture of Gaussians conditioned on the intermediate representation $Z_i$, which is obtained by passing the input sample $X_i$ through a feed-forward layer. 
Every input sample is thus associated with its own mixture of Gaussians.
Let $K$ denote the total number of mixtures in the MDN. Each component $k\in \{1, \cdots,K \}$ in the mixture is associated with a mixture coefficient $\rho_k$, mean $\mu_k$ and standard deviation $\sigma_k$.

Next, we provide a mathematical description of the proposed MDN models, specifically, MDN-C1. The input sample is first passed through a feed-forward layer with $r$ units. 
The output of the feed-forward layer for input $X_i$ is given by
\begin{equation}\label{eq:latent_eq}
\begin{aligned}
    Z_i=\mathrm{Tanh}(W_f \cdot X_i^T+b_f),   
\end{aligned}    
\end{equation}
where $\mathrm{Tanh}$ denotes the hyperbolic tangent activation, 
and $W_{f}$ is the $(r \times d)$-dimensional weight matrix of the feed-forward layer.
The intermediate representation $Z_i$ is then fed to the subsequent MDN layer to learn the distribution parameters. 
The parameters of the MDN with $K$ Gaussian components are obtained as 

\begin{equation}\label{eq:mdn_eq}
\begin{aligned}
\rho_{1,i},\cdots,\rho_{K,i}=\softmax(W_{\rho}\cdot Z_i^T+b_{\rho}), \\
\mu_{1,i},\cdots,\mu_{K,i}=W_{\mu}\cdot Z_i^T+b_{\mu},\\
\sigma_{1,i}, \cdots, \sigma_{K,i}=\softplus (W_{\sigma}\cdot Z_i^T+b_{\sigma}),
\end{aligned}
\end{equation}
where $W_{\rho}, W_{\mu}, W_{\sigma},b_{\rho}, b_{\mu}, b_{\sigma}$ represent the learnable parameters of the MDN layer with $\rho_{k,i}$, $\mu_{k,i}$ and $\sigma_{k,i}$ denoting the mixture weight, mean and standard deviation of the $k$th Gaussian component for the $i$th sample, respectively. 
The softmax activation ensures that $\rho_{k} \in [0,1]$ and $\sum_{k=1}^K\rho_{k} =1$, while the softplus activation ensures that the standard deviation term $\sigma$ is positive. 
The outputs of the MDN as formulated in (\ref{eq:mdn_eq}) model the conditional distribution of the class labels $\hat{Y}_1,\cdots,\hat{Y}_C$ by leveraging the universal approximation property of mixture of Gaussians. 
The weighted cumulative distribution function values are then evaluated using MDN parameters with each of the input dimension $X_{1,i},\cdots,X_{d,i}$ and each of the mixture component $k \in \bigr\{1, \cdots, K\bigr\}$ as the following:

\begin{equation}\label{eq:cdf_eq}
\begin{aligned}
\hat{\Phi}^{\mathrm{w}}_{k,m, i}=\rho_{k} \cdot {\Phi} \bigg( \frac{X_{m,i}-\mu_{k}}{\sigma_{k}} \bigg), 
\end{aligned}
\end{equation}
where ${\Phi}$ is the standard normal CDF and $\hat{\Phi}^{\rm w}_{k,m,i}$ is the weighted CDF value of the $k$th mixture component evaluated on the $i$th sample of the $m$th input dimension. It is convenient to view $\hat{\Phi}^{\rm w}_{k,m,i}$ as the $(m,k)$th element of a $d\times K$ dimensional vector which, through a slight abuse of notation, we denote by $\hat{\Phi}^{\rm w}_{i}$.
The weighted CDF values $\hat{\Phi}^{\rm w}_{i}$ are then passed through a feed-forward layer with softmax activation function to get the class scores as 
\begin{equation}\label{eq:preds_eq}
\begin{aligned}
\hat{Y}_{1,i},\cdots,\hat{Y}_{C,i}=\softmax(W_{o}\cdot (\hat{\Phi}_i^{\mathrm{w}})^{\rm T}+b_{o}),
\end{aligned}
\end{equation}
where $W_{o}, b_{o}$ are the learnable parameters of the softmax layer and $W_{o}$ is $C \times dK$.
The model parameters are learnt by optimizing the cross-entropy loss between the ground truth class labels and the predicted class scores given by 

\begin{equation}\label{eq:CE_loss}
\begin{aligned}
\mathcal{L}_{CE} = - \displaystyle \frac{1}{N} \displaystyle \sum_{i=1}^{N}
\sum_{c=1}^{C} Y_{i,c}\cdot \log \hat{Y}_{i,c},
\end{aligned}
\end{equation}
where $c \in \{1, \cdots, C\}$ is a class, 
$i$ denotes the $i$th input sample and $N$ is the total number of samples in the train set. 
The LASSO penalty~\cite{gugulothu2018sparse,gugulothu2019load,gugulothu2019sparse,tibshirani1996regression} is imposed on the weights $W_o$, so that only the CDF values of the most important mixture component's are used to predict the class scores. 
The final loss function along with the LASSO penalty ($L_1$ constraint) on the weights of the final softmax layer is thus given by

\begin{equation}\label{eq:sparse_negative_log_likelihood}
\begin{aligned}
\mathcal{L}_{} = \mathcal{L}_{CE} +\lambda \cdot \|{W}_o\|_1.
\end{aligned}
\end{equation}
The regularization parameter $\lambda$ controls the sparsity level in $W_o$ and helps to ensure that only the important CDF values are used to predict the class scores.

\begin{table*}[!ht]
\resizebox{\textwidth}{!}{
\begin{tabular}{|c|c|c|c|c|c|}
\hline
{\bfseries Dataset} & {\bfseries Model } & {\bfseries Accuracy } & {\bfseries Precision } & {\bfseries Recall } & {\bfseries F1-score } \\ 
\hline 
\hline
\multirow{7}{*}{\bfseries Pima Indians Diabetes }  
& Gaussian NB&0.764(0.023)&0.762(0.021)&0.764(0.023)&0.760(0.020)\\
\cline{2-6}
& Random Forest&0.774(0.024)&0.770(0.025)&0.774(0.024)&\textbf{0.769}(0.025)\\
\cline{2-6}
& SVM&\textbf{0.776}(0.017)&\textbf{0.775}(0.021)&\textbf{0.776}(0.017)&0.764(0.014)\\
\cline{2-6}
& XGBoost&0.756(0.017)&0.752(0.019)&0.756(0.017)&0.749(0.017)\\
\cline{2-6}
&ANN&0.765(0.018)&0.761(0.020)&0.765(0.018)&0.759(0.018)\\
\cline{2-6}
&MDN-C1&{0.769}({0.010})&{0.771}({0.014})&{0.769}({0.010})&\textbf{0.769}({0.011})\\
\cline{2-6}
&MDN-C2&0.768(0.013)&0.764(0.014)&0.768(0.013)&0.764(0.014))\\
\hline 
\hline
\multirow{7}{*}{\bfseries Waveform Generator } 
&Gaussian NB&0.795(0.009)&0.831(0.006)&0.795(0.009)&0.782(0.010)\\
\cline{2-6}
&Random Forest&0.848(0.006)&0.850(0.006)&0.848(0.006)&0.847(0.006)\\
\cline{2-6}
&SVM&0.857(0.007)&0.858(0.007)&0.857(0.007)&0.857(0.007)\\
\cline{2-6}
&XGBoost&0.844(0.008)&0.845(0.007)&0.844(0.008)&0.844(0.008)\\
\cline{2-6}
&ANN&0.860(0.011)&0.861(0.011)&0.860(0.011)&0.859(0.012)\\
\cline{2-6}
&MDN-C1&0.859(0.009)&0.859(0.009)&0.859(0.009)&0.858(0.009)\\
\cline{2-6}
&MDN-C2&\textbf{0.862}({0.008})&\textbf{0.863}({0.008})&\textbf{0.862}({0.008})&\textbf{0.861}({0.009})\\
\hline 
\hline
\multirow{7}{*}{\bfseries Multiple Features} 
&Gaussian NB&0.963({0.006})&0.964({0.006})&0.963({0.006})&0.963({0.006})\\
\cline{2-6}
&Random Forest&\textbf{0.988}({0.004})&0.988({0.004})&\textbf{0.988}({0.004})&0.987({0.004})\\
\cline{2-6}
&SVM&\textbf{0.988}({0.003})&\textbf{0.989}({0.003})&\textbf{0.988}({0.003})&\textbf{0.988}({0.003})\\
\cline{2-6}
&XGBoost&0.979({0.004})&0.980({0.004})&0.979({0.004})&0.979({0.004})\\
\cline{2-6}
&ANN&{0.987}({0.002})&{0.987}({0.002})&{0.987}({0.002})&{0.987}({0.002})\\
\cline{2-6}
&MDN-C1&{0.987}({0.003})&{0.987}({0.003})&{0.987}({0.003})&0.986(0.003)\\
\cline{2-6}
&MDN-C2&0.985(0.004)&0.985(0.004)&0.985(0.004)&0.984(0.004)\\
\hline
\end{tabular}}
\caption{Performance comparison of the proposed MDN-based classification models. The numbers in parentheses represent standard deviation values.}\label{tab:results}
\end{table*}

The second variant of our proposed MDN-based classification model, MDN-C2 (Figure~\ref{fig:MDN-Var2}) differs from MDN-C1 in one chief aspect. 
In MDN-C2, the input samples are passed through a feed-forward layer having as many units as the total number of classes ($C$) in the dataset, to learn the latent features by reducing the original dimensions $d$ to $C$. The objective of this feedforward layer is to cut down the number of CDF evaluations required to get the final class scores by a factor $\frac{d}{C}$ times, which is particularly beneficial in scenarios with very high input dimensions.
The latent features $f_{1,i}, \ldots, f_{C,i}$ are used to evaluate the CDF values as in (\ref{eq:cdf_eq}) in place of the original features $X_{1,i}, \ldots, X_{d,i}$. 
The MDN parameters are used to evaluate the CDF value $\hat{\Phi}_{m,i}$ corresponding to the $i$th sample of the $m$th latent feature as
\begin{equation}\label{eq:cdf_eq_var2}
\begin{aligned}
\hat{\Phi}_{m,i}= \displaystyle \sum_{k=1}^{K} \hat{\Phi}^{\mathrm{w}}_{k,m,i}, 
\end{aligned}
\end{equation}
where $m \in \{1,\ldots,C \}$. The evaluated CDF values ${\hat{\Phi}_{1,i}, \ldots, \hat{\Phi}_{C,i} }$ are then normalized to get the class scores in the range $[0, 1]$. Thus, the normalized class score on a sample $i$ for a class $c \in \{1,\ldots,C \}$ is given by
\begin{equation}\label{eq:cdf_eq_var_softmax}
\begin{aligned}
\hat{Y}{c,i}= {\left[\displaystyle \sum_{c=1}^{C} \hat{\Phi}_{c,i}\right]}^{-1}\hat{\Phi}_{c,i}.
\end{aligned}
\end{equation}
The predicted class scores are then used to calculate the cross-entropy loss as in (\ref{eq:CE_loss}) for MDN-C2. 

\section{Performance Evaluation}\label{sec:experiments} 
In this section, we compare the performance of the proposed MDN-based models for classification 
with five popular classification models, 
namely, Gaussian naive Bayes (NB), random forest, support vector machine (SVM), XGBoost and ANN.
We ran experiments on three publicly available datasets from UCI machine learning data repository\footnote{https://archive.ics.uci.edu/ml/datasets.php}
~\cite{Dua:2019}, specifically, i) the Pima Indians Diabetes dataset, ii) the Waveform Generator dataset, and iii) the Multiple Features dataset. 
We considered accuracy, precision, recall and F1-score metrics to measure the performance of the proposed MDN models.

\begin{table}[h]
\resizebox{\columnwidth}{!}{
\begin{tabular}{|c|c|c|c|c|}
\hline
{Dataset}&{Classes ($C$)}&{Input dimensions ($d$)}&{Total samples}&Class split(\%)\\
\hline 
\hline
Pima Indians&2&8&768&65(Class-0)/35(Class-1)\\
\hline
Waveform&3&40&5000&3 equiprobable classes\\
\hline
Multiple Features&10&649&2000&10 equiprobable classes\\ 
\hline
\end{tabular}}
\caption{Datasets details.} \label{tab:datasets_description}
\end{table}

\subsection{Datasets Description}\label{sec:datasets}
We provide a description of the datasets used in this work below. 
\begin{enumerate}
    \item \textbf{Pima Indians Diabetes dataset:} The classification task on this dataset is to predict whether or not a patient has diabetes based on certain diagnostic measurements such as age, number of pregnancies, plasma glucose, blood pressure, insulin etc. 
    The diagnostic measurements are of females of Pima Indian heritage with age 21 years or more.
    \item \textbf{Waveform Generator dataset:} This dataset consists of three classes of waveforms. 
    Each class is generated from a combination of two of three base waves. 
    Each waveform is described by a total of 40 noisy features, 19 of which are all noise attributes with mean 0 and variance 1. 
    In this dataset, the task is to classify the input sample into one of the three classes.
    \item \textbf{Multiple Features dataset:} This dataset consists of features of hand written digits (0-9).
    There are 200 patterns for each digit. 
    Six features such as Fourier coefficients, profile correlations, Karhunen-Love coefficients, Zernike moments etc, are extracted from the character shapes to represent the digits. 
    The task in this dataset is to classify the sample into one of the ten classes. 
\end{enumerate}

 Table~\ref{tab:datasets_description} provides more details of the datasets including the number of classes $C$, theinput dimensions $d$, total samples and class split.
 As Table~\ref{tab:datasets_description} shows, the datasets considered in this exposition have sufficient variety with the number of classes ranging from 2 to 10 and the input dimensions varying from 8 to 649.

\subsection{Training Details}\label{sec:training}
In our training process, each dataset was divided into train, validation
and test sets.
We used z-normalization to normalize the train, validation
and test sets by obtaining train set statistics.
We also employed the $k$-fold cross-validation technique with $k$ being $5$ in the performance evaluation of the models.

The optimal neural network is selected as the one with 
 the least cross-entropy loss on the hold-out validation set in each fold via a grid search on following hyper-parameters: feed-forward layers in $Z$ are $l\in\{1,2\}$, the number of units per layer $r\in\{5,\ldots,100\}$,
 the number of mixtures in MDN layer $K\in\{1, 2, 3, 4, 5, 10\}$ and a dropout rate of $0.3$ in the feed-forward layers for regularization.
The sparsity controlling parameter $\lambda$ is chosen from $\{0, 0.001, 0.0001\}$. 

A similar tuning procedure is
used to optimize the performance of the ANN based classification model. 
In random forest, 
the number of estimators $e \in \{5,\ldots,300\}$ are tuned. 
We used SVM with radial basis function kernel and tuned $c\in \{0.001, 1, 10, 100\}$ and $\gamma \in \{0.001, 1, 10, 100\}$ parameters. XGBoost is trained with softmax objective and the following hyper-parameters are tuned: 
the number of estimators $e \in \{5,\ldots,300\}$,
$\gamma \in \{0.001, 1, 10, 100\}$, learning rate $\eta \in \{0.01,0.1,0.2\}$.
As discussed above, each experiment is run for five folds. 
The mean and standard deviation values are reported in Table~\ref{tab:results}.

\subsection{Results and Observations} \label{sec:results}
The performance of proposed MDN-based classification models are summarized in Table~\ref{tab:results}. We make the following observations from the results: 
\begin{itemize}
    \item The proposed MDN-C models are marginally superior to the Gaussian naive Bayes, XGBoost and ANN based classification models on all metrics and at par with the other baselines for the Pima Indians Diabetes dataset. 
    \item On the Waverform Generator dataset, the proposed MDN-C2 performed better than all the baselines on all metrics. 
    \item The proposed MDN-C models are slightly better than Gaussian naive Bayes and XGBoost and performed at par with the other baselines on the Multiple Features dataset on all metrics. 
\end{itemize}

\begin{figure}[h]
\includegraphics[trim={0.5cm 4.0cm 11cm 0cm},clip,width=0.35\textwidth] {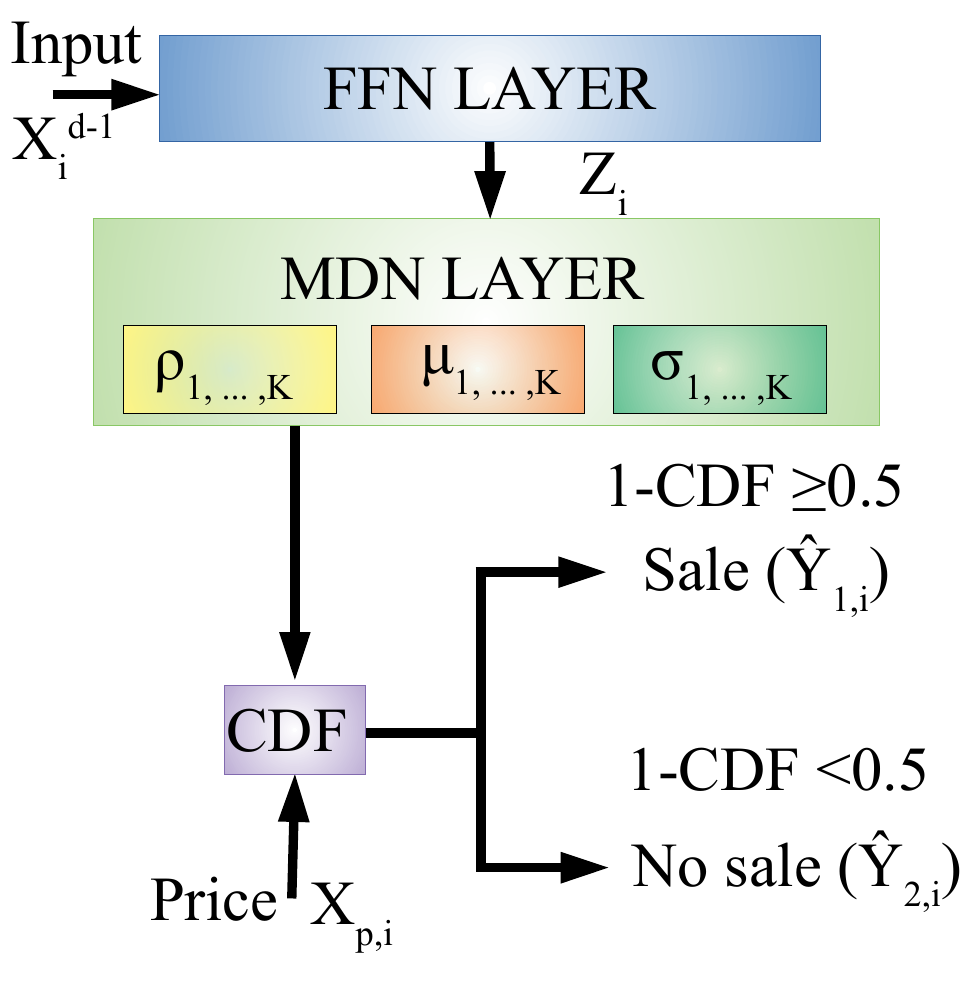} 
\captionof{figure}{Product sales classification using MDN-P2B.}\label{fig:MDN-C_bundling}
\end{figure}
\begin{figure*}[!h]
	\subfigure[\label{fig:purchprob-item1}Product-1 purchase probability $V_1(p)$]{\includegraphics[trim={0cm 0cm 0cm 0cm},clip,width=0.5\textwidth]{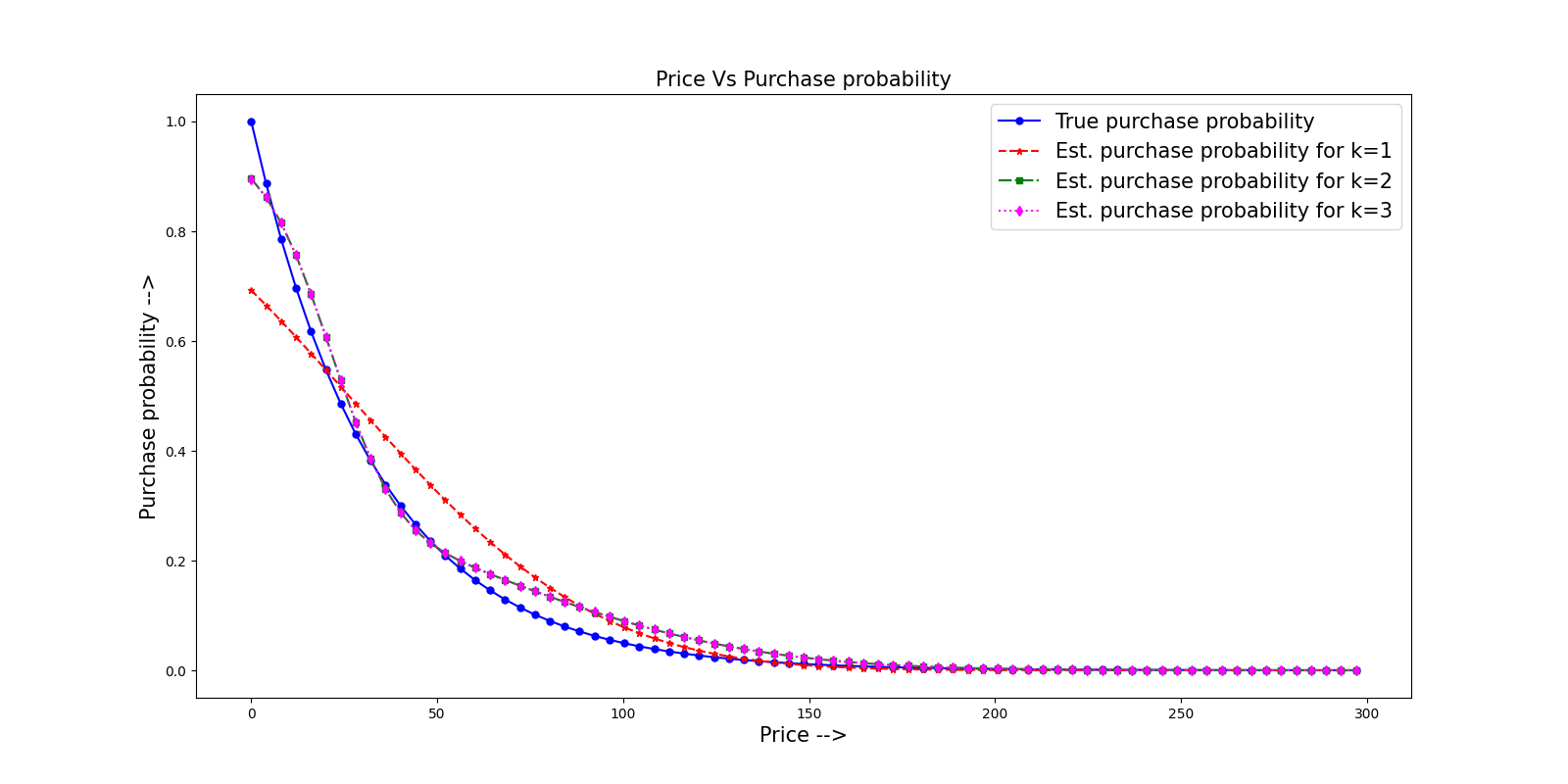}}
	\subfigure[\label{fig:purchprob-item2}Product-2 purchase probability $V_2(p)$ ]{\includegraphics[trim={0cm 0cm 0cm 0cm},clip,width=0.5\textwidth]{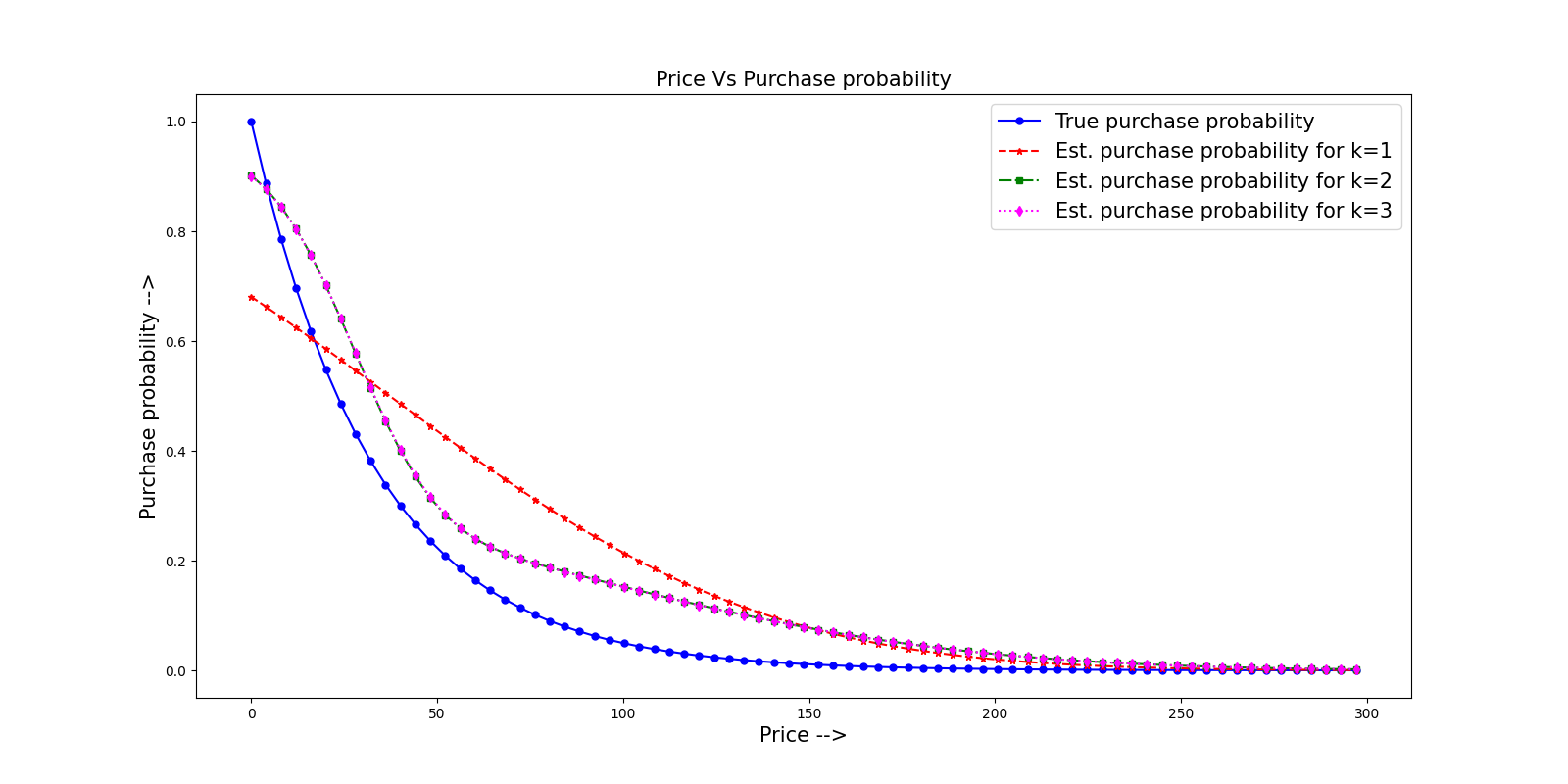}}
    \subfigure[\label{fig:purchprob-bundle}Bundle purchase probability $V_{\rm b}(p)$  ]{\includegraphics[trim={0cm 0cm 0cm 0cm},clip,width=0.5\textwidth]{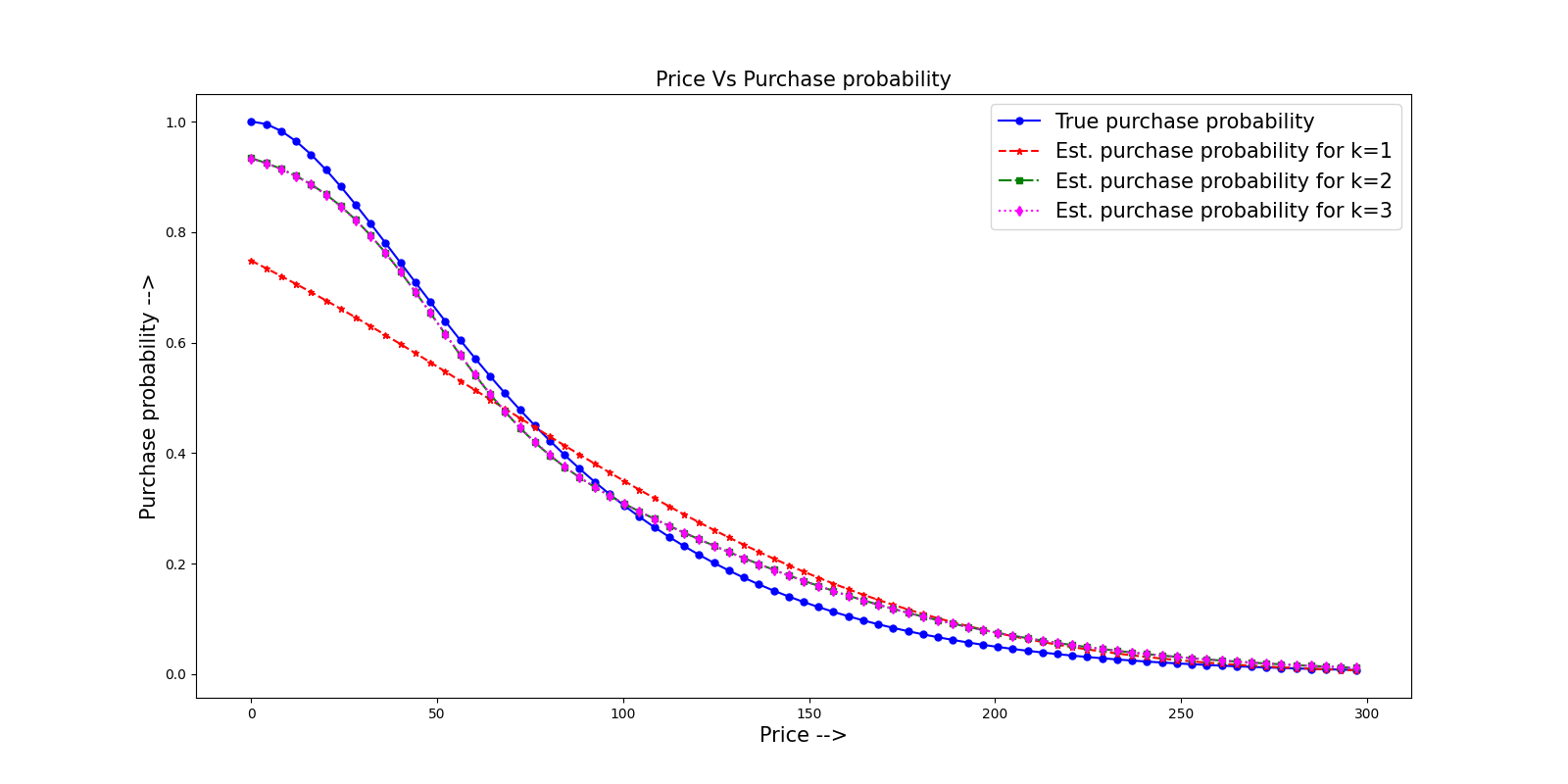}}
    \subfigure[\label{fig:rev-bundle}Bundle revenue]{\includegraphics[trim={0cm 0cm 0cm 0cm},clip,width=0.5\textwidth]{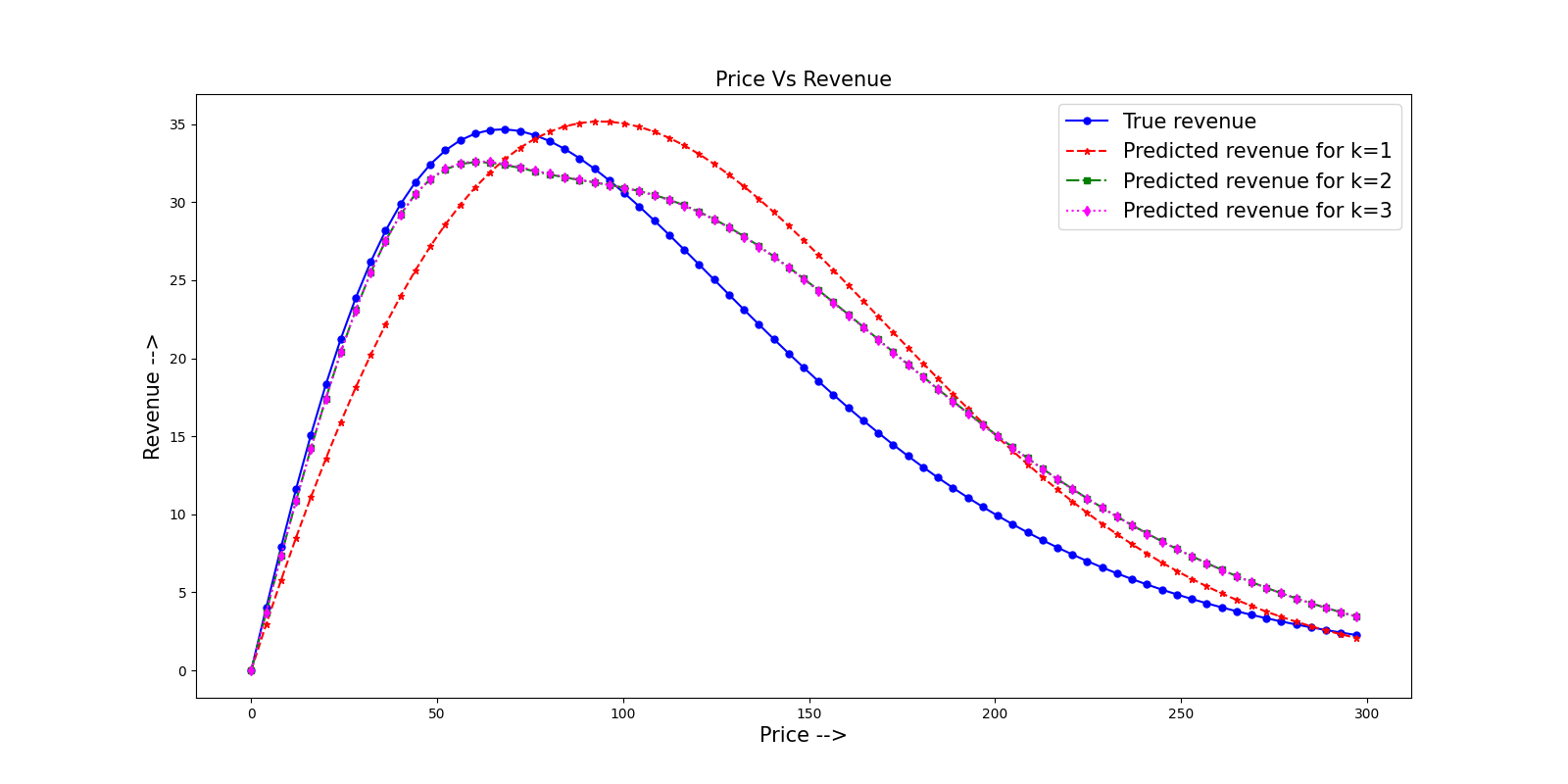}}
	\caption{\label{fig:bundling} Product bundling results using MDN-P2B. Best viewed on zooming.}
\end{figure*}

\section{Product Bundling}\label{sec:bundling}
In this section, we consider an application of the proposed MDN-based models to product bundling. Our objective is to estimate the WTP distribution for
a bundle of products from the WTP distributions for individual products, learnt from individual products sales data using the proposed MDN-C1 model.
As the ground truth model, we assume that the 
WTP distribution for products is exponential with rate $\lambda$. 
This ground truth model is used to generate product-level sales data as follows: each arriving customer is offered a product $i$ at a price $p$. At the same time, the customer's WTP is sampled from the assumed WTP distribution, and a sale is recorded if and only if the sampled WTP is greater than the offer price $p$. The data consists only of the offer price and a binary indicator indicating if a sale was made. 
Note that the underlying WTP distribution is typically unknown. 
Towards this, we use the proposed MDN-C1 model with minor modifications to approximate the product-level WTP distributions by using the sales data. The underlying classification task is to predict sale or no sale given the price. 
The learnt product-level WTP distribution parameters, namely, mixture coefficients $\rho_{1,\cdots,K}$, means $\mu_{1,\cdots,K}$ and standard deviations $\sigma_{1,\cdots,K}$ are then used to estimate the bundle WTP distribution via convolution as described below.

In the rest of this section, we consider a bundle with two products $i_1$ and $i_2$ and denote the corresponding WTP random variable by $T_1$ and $T_2$ respectively.
We assume that the corresponding WTP distribution for products $i_1$ and $i_2$ follow exponential distributions with rates $\lambda_1=0.03$ and $\lambda_2 = 0.02$ respectively. 
Then the purchase probability $V_1(p):= P(T_1 \geq p)$ for product $i_1$ is equal to $\exp{(-\lambda_1\cdot p)}$.
Similarly, for product $i_2$, we have $ V_2(p)=\exp{(-\lambda_2\cdot p)}$. Central to our work is the assumption that the bundle WTP variable $T_{\rm b}$ is an addition of the WTP variables $T_1$ and $T_2$, i.e., $T_{\rm b} = T_1 + T_2$. 
The assumption is a standard one in choice modeling theory, and is based on the intuition that a customer will buy a bundle if and 
only if it is offered at a price that does not exceed the sum of the individual amounts that he is willing to pay for the products separately. 
Since the WTP for the individual products are independent, the WTP distribution of the bundle is easily seen to be the convolution of the WTP distributions of the individual products. 
Thus, under our assumption, it can be seen that $T_{\rm b}$ follows a hypo-exponential distribution with parameters $\lambda_1$ and $\lambda_2$.
Furthermore, it's mean value is $\frac{1}{\lambda_1}+\frac{1}{\lambda_2}$ and the corresponding bundle purchase probability $V_{\rm b}(p)$ at price $p$ equals $\frac{\lambda_2}{\lambda_2-\lambda_1} \exp(-\lambda_1 \cdot p) - \frac{\lambda_1}{\lambda_2-\lambda_1} \exp(-\lambda_2 \cdot p)$. 

In this experiment, we used MDN-C1 model with following two modifications
and call it as MDN-Products2Bundle (MDN-P2B): i) input to the intermediate term $Z$ is dummy and constant, so that the MDN parameters are independent of the offered price, ii) The final softmax layer, which predicts class scores, is discarded and the CDF value (as in~\ref{eq:cdf_eq_var2}), evaluated with the offered price, is used to predict class scores as the CDF value is sufficient for binary classification as shown in Figure~\ref{fig:MDN-C_bundling}. 
Let the learnt product-level WTP probability density functions (PDFs) using the MDN-P2B model be $f(p)=\displaystyle \sum_{k=1}^K \rho_{k} \mathcal{N}(p \, | \, \mu_{k},\sigma_{k})$, and $g(p)=\displaystyle \sum_{j=1}^M \rho_{j} \mathcal{N}(p \, | \, \mu_{j},\sigma_{j})$ for products $i_1$ and $i_2$, respectively. 
Since $T_{\rm b} = T_1 + T_2$, the PDF for $T_{\rm b}$ can be obtained as the convolution $h(\cdot)=(f\ast g)(\cdot)$ of $f(\cdot)$ and $g(\cdot)$. This convolution is particularly straightforward since $f$ and $g$ are mixtures of Gaussians, and is given by
\begin{equation}\label{eq:conv}
\begin{aligned}
h(p)&=\left(\displaystyle \sum_{k=1}^K \rho_{k} \mathcal{N}(p \, | \, \mu_{k},\sigma_{k})\right) \ast \displaystyle \left(\sum_{j=1}^M \rho_{j} \mathcal{N}(p \, | \, \mu_{j},\sigma_{j})\right) \\
&= \displaystyle \sum_{k=1}^K \displaystyle \sum_{j=1}^M \rho_{k} \rho_{j} \, \mathcal{N}\Big(p \, | \, \mu_{k}+\mu_{j},\sqrt{\sigma_{k}^2+\sigma_{j}^2}\Big).
\end{aligned}
\end{equation}


Note that, though we considered a bundle with two products in this exposition, the ideas can be easily extended to more products.
In our experiments, the offered price for the products and bundle was in the range of 0-300. The optimal network was obtained by tuning the following hyper-parameters:
units $r \in \{5,\ldots ,100\} $, layers $l \in \{1,2\}$ and the number of mixture components $K\in\{1, 2, 3\}$. We ran the experiments for five seeds and the performance of the MDN-P2B for product-level sales classification is reported in Table~\ref{tab:item_class} for $K=3$. 
The average and standard deviation (in parentheses) of the estimated mean of the learned WTP distribution of five runs are reported in Table~\ref{tab:bundling}. 
\begin{table}[th]
\resizebox{\columnwidth}{!}{
\begin{tabular}{|c|c|c|}
\hline
{\bfseries Metric} & {\bfseries Product-1 } & {\bfseries Product-2 }  \\ 
\hline \hline
{Accuracy}&0.917 (0.007)&0.878 (0.006)\\
\hline
{Precision}&0.911 (0.008)&0.874 (0.004) \\
\hline
{Recall}&0.917 (0.007)&0.878 (0.006)\\
\hline
{F1-score}& 0.909 (0.009)&0.864 (0.014)\\
\hline
\end{tabular} 
}
\caption{Results on product-level sales classification using MDN-P2B model.\label{tab:item_class}}
\end{table}
\begin{table}[th]
\resizebox{\columnwidth}{!}{
\begin{tabular}{|c|c|c|c|c|}
\hline
{\bfseries }&{\bfseries True Mean}&\multicolumn{3}{c|}{\bfseries Estimated Mean}\\
\cline{3-5}
&&$K=1$&$K=2$&$K=3$\\
\hline \hline
{Product-1}&33.33&25.24(0.76)&32.80(1.56)&\textbf{33.00}(1.54) \\
\hline
{Product-2}&50&36.93(0.65)&48.83(1.80)&\textbf{49.51}(1.69) \\
\hline
{Bundle}&83.33&62.17(1.09)&81.63(2.63)&\textbf{82.51}(2.27) \\
\hline
\end{tabular}
}
\caption{Results on product bundling using MDN-P2B.} \label{tab:bundling}
\end{table}
The estimated purchase probabilities for both the products and the bundle as well as the corresponding true probabilities calculated according to the ground truth model are plotted as functions of the offer price $p$ in Figures~\ref{fig:purchprob-item1}, \ref{fig:purchprob-item2} and \ref{fig:purchprob-bundle}. 
The estimated revenue for bundle is shown in Figure~\ref{fig:rev-bundle}.

We make following observations from Table~\ref{tab:bundling} and Figure~\ref{fig:bundling}:
\begin{itemize}
    \item Product-level WTP distributions are learnt well as shown in Figures~\ref{fig:purchprob-item1} and~\ref{fig:purchprob-item2}. The true and estimated purchase probabilities are matching. True and estimated mean parameters are also very close for $K=3$.
    \item Increasing the number of mixture components helps
    to learn the true WTP distributions better. Also, MDN-P2B
    with $K=3$ performs better.
    \item The approximated 
    purchase probability and expected revenue for the bundle closely match the ground truth as shown in figures ~\ref{fig:purchprob-bundle} and ~\ref{fig:rev-bundle}. 
    \item The learnt bundle distribution is then used to estimate revenue maximizing price (Figure~\ref{fig:rev-bundle}).     
\end{itemize}

\section{Discussion}\label{sec:discussion}
We proposed two variants of MDN-based models namely, MDN-C1 and MDN-C2 for classification. 
These models learn the underlying distribution parameters.
The performance of the proposed MDN-based models was compared with five classification baselines on three publicly available datasets. 
Apart from performing better or on par, these models can be employed in classification tasks where learning the distribution parameters is crucial. 
The usefulness of learned distribution parameters is illustrated by considering a product bundling problem.
Also, the proposed MDN-C models can be combined and trained in an end-to-end manner with deep-learning techniques like RNNs, CNNs and transformer networks. 
In future, it would be interesting to use mixture distributions of Gamma, Levy and exponential etc., to learn bundle WTP distributions from product-level data and also to learn product-level WTP distributions using bundle-level sales data.

\bibliography{Class_AAAI2024}

\end{document}